\documentclass[11pt,a4paper]{article}
\usepackage[hyperref]{acl2021}
\usepackage{times}
\usepackage{latexsym}

\usepackage{stix}
\usepackage{colortbl}
\usepackage{siunitx}
\usepackage{xurl}
\usepackage{siunitx}
\usepackage{leftidx}
\usepackage{booktabs}
\usepackage{algorithm}
\usepackage{arydshln}
\usepackage[noend]{algpseudocode}
\usepackage{amsmath}
\usepackage{float}
\usepackage{multirow}
\usepackage{subcaption}
\usepackage{mathtools}
\usepackage{enumitem}

\usepackage{url}
\usepackage{cleveref}
\crefformat{section}{\S#2#1#3}
\crefformat{subsection}{\S#2#1#3}
\crefformat{subsubsection}{\S#2#1#3}
\crefrangeformat{section}{\S\S#3#1#4 to~#5#2#6}
\crefmultiformat{section}{\S\S#2#1#3}{ and~#2#1#3}{, #2#1#3}{ and~#2#1#3}

\definecolor{nigel}{RGB}{255, 40, 40}
\definecolor{yingzhen}{RGB}{0, 153, 51}
\definecolor{ehsan}{RGB}{0, 40, 255}
\definecolor{victor}{RGB}{255, 140, 0}

\usepackage{microtype}

\aclfinalcopy 

\title{Learning Sparse Sentence Encoding without Supervision:\\ An Exploration of Sparsity in Variational Autoencoders}
\author{
  Victor Prokhorov$^\clubsuit$
  \ \ Yingzhen Li$^\diamondsuit$\thanks{\enspace Work done while at Microsoft Research Cambridge.}
  \ \ Ehsan Shareghi$^\spadesuit$$^\clubsuit$

  \ \ Nigel Collier$^\clubsuit$\\
  $^\clubsuit$~Language Technology Lab, University of Cambridge\\
  $^\spadesuit$~Department of Data Science \& AI, Monash University \\
  $^\diamondsuit$~Department of Computing, Imperial College London\\
  
  {\tt vp361@cam.ac.uk},  {\tt yingzhen.li@imperial.ac.uk}, \\ {\tt ehsan.shareghi@monash.edu},
  {\tt nhc30@cam.ac.uk}
}

\date{}

\begin{document}
\maketitle
\begin{abstract}
 It has been long known that sparsity is an effective inductive bias for learning efficient representation of data in vectors with  \emph{fixed dimensionality}, and it has been explored in many areas of representation learning.  Of particular interest to this work is the investigation of the sparsity within the VAE framework which has been explored a lot in the image domain, but has been lacking even a basic level of exploration in NLP. Additionally, NLP is also lagging behind in terms of learning sparse representations of large units of text e.g., sentences. We use the VAEs that induce sparse latent representations of large units of text to address the aforementioned shortcomings.  First, we move in this direction by measuring the success of unsupervised state-of-the-art (SOTA) and other strong VAE-based sparsification baselines for text and propose a hierarchical sparse VAE model to address the stability issue of SOTA. Then, we look at the implications of sparsity on text classification across 3 datasets, and highlight a link between performance of sparse latent representations on downstream tasks and its ability to encode task-related information.
\end{abstract}

\section{Introduction}
Representation learning has been pivotal in many success stories of modern days NLP. 
%
Observing its success, two fundamental questions arise: \emph{How the information is encoded in them?} and \emph{What is encoded in them?} While the latter has received a lot of attention by designing probing tasks, the former has been vastly neglected. In this work, we take small steps in this non-trivial direction by building on the knowns: One property we know about the encoding of information is that different data points embody different characteristics (e.g. statistically, semantically, or syntactically) which should ideally utilise different sub-regions of the representation space. Therefore, the high-dimensional learned representations should ideally be sparse~\cite{DBLP:journals/pami/BengioCV13,DBLP:journals/corr/abs-1804-03599,tonolini2019vsc}. But \emph{if sparsity\footnote{As in \cite{pmlr-v97-mathieu19a}, we induce sparse representations for each data point.} is expected, could it be learned from data without supervision?}

A handful of studies in NLP that have delved into building sparse representations of words either during the learning phase~\cite{faruqui-dyer-2015-non, DBLP:conf/icml/YogatamaFDS15} or as a post-processing step on top of existing representations (e.g., word2vec embeddings)~\cite{faruqui-etal-2015-sparse,DBLP:conf/ijcai/SunGLXC16, DBLP:conf/aaai/SubramanianPJBH18, arora-etal-2018-linear, DBLP:journals/corr/abs-1911-01625}. These methods have not been developed for sentence embeddings, with the exception of 
\citet{DBLP:conf/emnlp/TrifonovGPH18} 
which makes a strong  assumption by forcing the latent sentence representation to be a sparse categorical distribution.

In parallel, Variational Autoencoders~(VAEs)~\cite{DBLP:journals/corr/KingmaW13} have been effective in capturing semantic closeness of sentences in the learned representation space~\cite{DBLP:journals/corr/BowmanVVDJB15,
prokhorov-etal-2019-importance, xu2019variational,balasubramanian2020polarized}. Furthermore, methods have been developed to encourage sparsity in VAEs via learning a deterministic selection variable~\cite{Yeung2017TacklingOI} or sparse priors~\cite{ Barello399246, pmlr-v97-mathieu19a,tonolini2019vsc}. However, the success of these is yet to be examined on text domain.

To bridge this gap, we make a sober evaluation of existing state-of-the-art (SOTA) VAE-based sparsification model~\cite{pmlr-v97-mathieu19a} against several VAE-based baselines on two experimental tasks: text classification accuracy, and the level of representation sparsity achieved. Additionally, we propose Hierarchical Sparse Variation Autoencoder~(HSVAE), to improve the stability issue of existing SOTA model and demonstrate how its favorable flexibility on both experimental tasks. 

Our experimental findings demonstrate that: (I) neither the simpler baseline models nor the SOTA manage to impose a satisfactory level of sparsity on text, (II) as expected, sparsity level and task performance have a negative correlation, while giving up task performance and having sparse codes helps with the analysis of the representations, (III) presence/absence of task related signal in the sparsity codes affects the task performance, (IV) the success of capturing the task related signal in the sparsity codes depends on the strength of the signal presented in a corpus, and representation dimensionality, (V) the success of SOTA in image domain does not necessarily transfer to inducing sparse representations for text, while HSVAE addresses this shortcoming.


\section{Background} 
\paragraph{VAE.} 
Given an input $x$, VAEs, Figure~\ref{figure:graph_model}~(left), are stochastic autoencoders that map $x$ to a corresponding representation $z$ using a probabilistic encoder $q_\phi(z|x)$ and a probabilistic decoder $p_\theta(x|z)$, implemented as neural
networks. Optimisation of VAE is done by maximising the ELBO:
{
\setlength{\abovedisplayskip}{3pt}
\setlength{\belowdisplayskip}{3pt}
\setlength{\abovedisplayshortskip}{3pt}
\setlength{\belowdisplayshortskip}{3pt}
\begin{align}\label{elboobjective}
\mathbb{E}_{q_\phi({z}|{x})}\log p_\theta({x}|{z})   -  \vphantom{\big \langle \log p_\theta({x}|{z}) \big \rangle_{{z} \sim q_\phi({z}|{x})}} \mathbb{D}_{KL}\big(q_\phi({z}|{x}) || p_\theta({z})\big)
\end{align}
}
where the reconstruction maximises the expectation of data likelihood under the posterior distribution of $z$, and the Kullback-Leibler (KL) divergence acts as a regulariser and minimises the distance between the learned posterior and prior of $z$. 
\paragraph{Spike-and-Slab Distribution.} 
This is a mixture of two Gaussians with mixture weight $\gamma_i$, where the \emph{slab} component is a standard Gaussian while the \emph{spike} component is a Gaussian with $\sigma \rightarrow 0$: 
{
\setlength{\abovedisplayskip}{3pt}
\setlength{\belowdisplayskip}{3pt}
\setlength{\abovedisplayshortskip}{3pt}
\setlength{\belowdisplayshortskip}{3pt}
\begin{align*}
p(z) = \prod_{i} (1-\gamma_i)\,\mathcal{N}(z_i;0,1)+ \gamma_i\,\mathcal{N}(z_i;0,\sigma \xrightarrow{\makebox[0.01cm]{}} 0)
\end{align*}
}
where $i$ denotes the $i$th dimension of $z$.

\section{Hierarchical Sparse VAE (HSVAE)}

We propose the hierarchical sparse VAE (HSVAE), Figure~\ref{figure:graph_model}~(right), to learn sparse latent codes automatically. We treat the mixture weights $\gamma = (\gamma_1, ..., \gamma_D)$ as a random variable and assign a factorised Beta prior $p_{\theta}(\gamma_i) = \text{Beta}(\alpha, \beta)$ on it. The latent code $z$ is then sampled from a factorised Spike-and-Slab distribution $p_{\theta}(z | \gamma)$ conditioned on $\gamma$, and the observation $x$ is generated by decoding the latent variable $x \sim p_{\theta}(x | z)$ using a GRU~\cite{DBLP:journals/corr/ChoMGBSB14} decoder. This returns a probabilistic generative model $p_{\theta}(x, z, \gamma) = p_{\theta}(x | z) p_{\theta}(z | \gamma) p_{\theta}(\gamma)$.

For posterior inference, the encoder distribution is defined as $q_{\phi}(z, \gamma | x) = q_{\phi}(\gamma | x) q_{\phi}(z | \gamma, x)$, where $q_{\phi}(\gamma | x)$ is a learnable and factorised Beta distribution, and $q_{\phi}(z | \gamma, x)$ is a factorised Spike-and-Slab distribution with mixture weights $\gamma_i$ and learnable ``slab'' components for each dimensions. 
The $q$ distribution is computed by first extracting features from the sequence using a GRU, then applying MLPs to the extracted feature (and $\gamma$ for $q_{\phi}(z|\gamma, x)$) to produce the distributional parameters.

\paragraph{ELBO:} We derive the ELBO, $\mathcal{L}(\theta, \phi; x)$:

\begin{equation*}\label{hsvae-objective}
\resizebox{0.93\hsize}{!}{$%
\begin{aligned}
 \mathbb{E}_{q_{\phi}(z, \gamma|x)}[\log p_{\theta}(x|z)] - \psi \mathbb{E}_{q_{\phi}(\gamma|x)}[ \mathbb{D}_{KL}\big(q_{\phi}(z | \gamma, x),\\p_{\theta}(z|\gamma)\big)]  - \lambda \mathbb{D}_{KL}\big(q_{\phi}(\gamma|x)||p_{\theta}(\gamma)\big),
\end{aligned}
$%
}
\end{equation*}

where $\psi \in \mathbb{R} $ and $\lambda \in \mathbb{R}$ are the coefficients for the KL terms. 
This ELBO is approximated with Monte Carlo (MC) in practice, $\mathcal{L}(\theta, \phi; x)$:
\begin{figure}[t]%
\centering
\includegraphics[scale=0.6,trim={0cm 0.66cm 0cm 0}, clip]{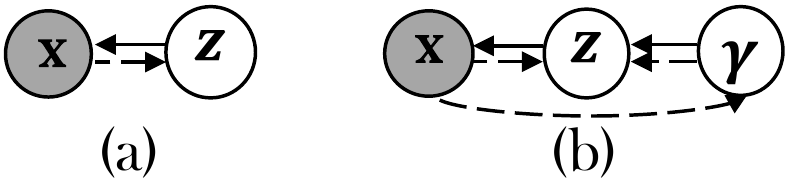}
\caption{ Graphical models of VAE~(left) and HSVAE~(right). Solid and dashed lines represent generative and inference paths, respectively.}
\label{figure:graph_model}
\end{figure}
\begin{equation}
\begin{split}
  \frac{1}{N}\sum_{\gamma \sim q_{\phi}(\gamma|x)}^N\bigg[ \frac{1}{M}\sum\limits_{z \sim q_{\phi}(z|x, \gamma)}^M \log p_{\theta}(x|z) \bigg]-\\
-\frac{\psi}{N}\sum_{\gamma \sim q_{\phi}(\gamma|x)}^N\bigg[ \mathbb{D}_{KL}(q_{\phi}(z|x, \mathbf{\gamma})||p_{\theta}(z|\gamma)) \bigg]-\\ -\lambda \mathbb{D}_{KL}(q_{\phi}(\gamma|x)||p_{\theta}(\gamma)),
\end{split}
\end{equation}
where $M$ and $N$ are scalar numbers corresponding to a number of samples taken from $q_{\phi}(z|x, \gamma)$ and $q_{\phi}(\gamma|x)$ respectively. In this work, we set both $M$ and $N$ to $1$.
Similar to the vanilla VAE, the first term is the reconstruction, the second and the third KL terms control the distance between the posteriors and their corresponding priors. See Appendix \ref{appendix:elbo_derivation} for ELBO derivation.
%
\paragraph{Control of Sparsity.} The random variable $\gamma_i$, in our model, can be viewed as a ``probabilistic switch'' that determines how likely is for the $i$th dimension of $z$ to be turned off. Intuitively,
since for both generation and inference the latent code $z$ is sampled from a Spike-and-Slab distribution with the mixture weights $\gamma$, 
$\gamma_i \rightarrow 1$ means $z_i$ is drawn from a delta mass centered at $z_i = 0$.
As the switch follows a Beta distribution $\gamma_i \sim Beta(\gamma_i;\alpha, \beta)$, we can select the parameters $\alpha$ and $\beta$ to control the concentration of the probability mass on $\gamma_i \in [0, 1]$ interval.

There are three typical configurations of the $(\alpha, \beta)$ pair: (1) $\alpha < \beta$: density is shifted towards $\gamma_i=0$ hence $i$th unit is likely to be on and dense representation is expected, (2) $\alpha = \beta$: the density is centered at $\gamma_i = 0.5$, and (3) $\alpha > \beta$: density is shifted towards $\gamma_i=1$, hence the unit is likely to be off, leading to sparsity. The magnitude of these parameters also plays a role as it controls the spread and uni/bi-modal structure of the density.

\section{Experiments}
\label{sec:experiment}
\begin{figure*}[t]
\centering
\begin{subfigure}[b]{1.0\textwidth}
   \includegraphics[trim=0 65 0 0,clip,width=1\linewidth]{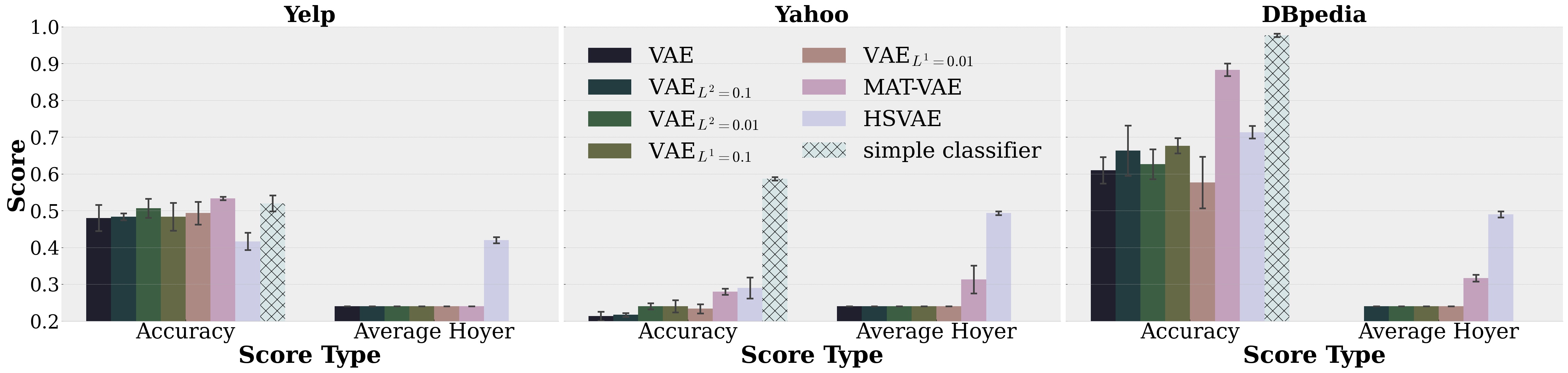}
   \caption{}
   \label{fig:Ng1} 
\end{subfigure}

\begin{subfigure}[b]{1.0\textwidth}
   \includegraphics[trim=0 65 0 0,clip,width=1\linewidth]{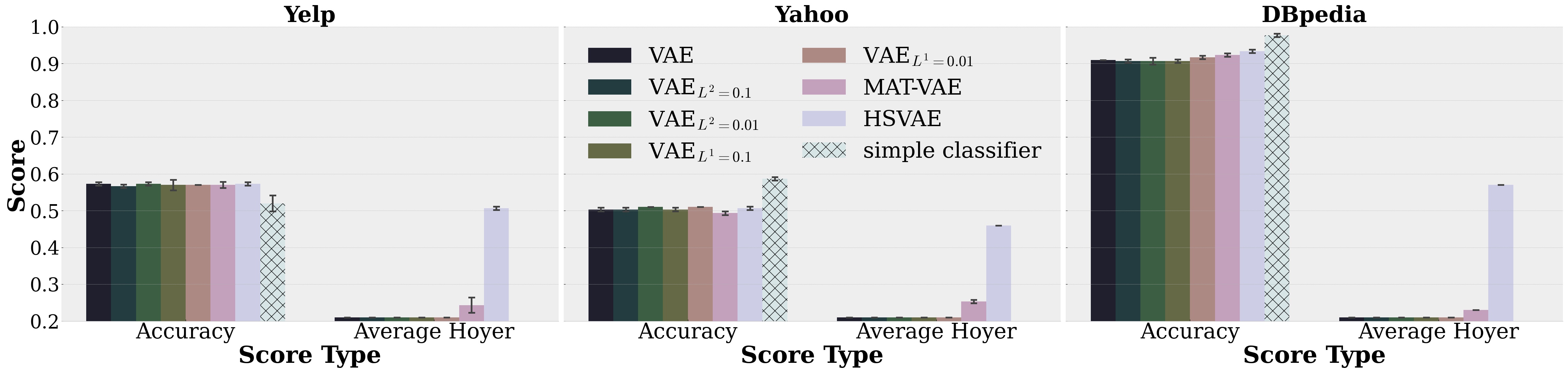}
   \caption{}
   \label{fig:Ng2}
\end{subfigure}

\begin{subfigure}[b]{1.0\textwidth}
   \includegraphics[trim=0 65 0 0,clip,width=1\linewidth]{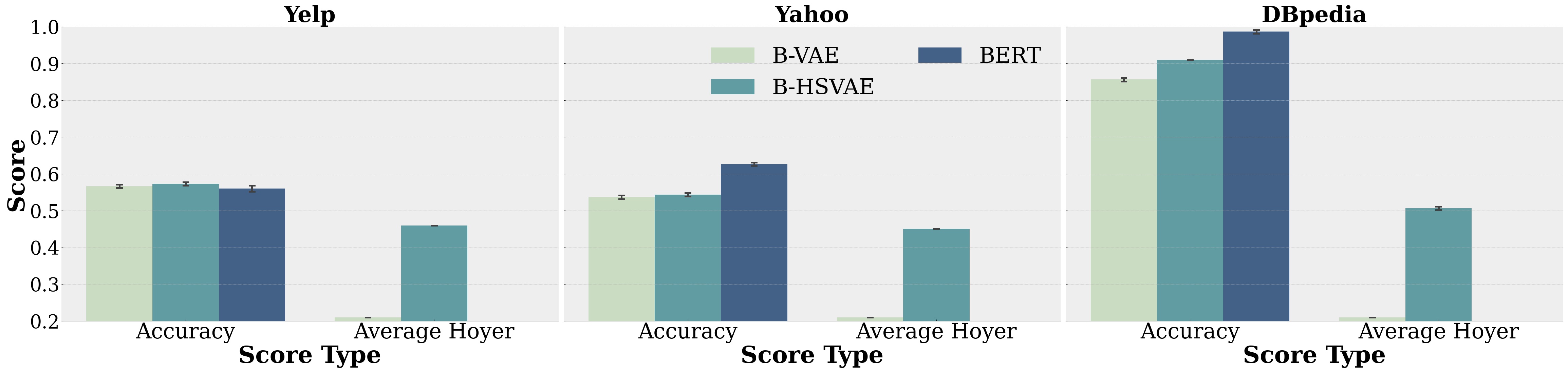}
   \caption{}
   \label{fig:Ng3}
\end{subfigure}
\vspace*{-6mm}
\caption[Two numerical solutions]{Classification Accuracy and Average Hoyer (higher means sparser $z$) for various VAE variants and the two baselines: simple classifier and BERT evaluated on Yelp, Yahoo or DBpedia test. The latent code of the VAEs is 32 D Figure (a) and 768 D Figures (b) and (c). Hoyer metric is not applicable to the simple classifier in the panels (a) and (b) and to the vanilla BERT model in the panel (c). The weights of the VAE encoders and BERT are \textbf{frozen} during the training of the classifiers. While the encoder of the simple classifier is updated during the training. }
\label{figure:acc_hoyer}
\end{figure*}

We conduct a set of  experiments on three text classification corpora: Yelp (sentiment analysis - 5 classes)~\cite{DBLP:journals/corr/YangHSB17}, DBpedia and Yahoo ~(topic classification - 14 and 10 classes respectively)~\cite{10.5555/2969239.2969312}. 
First, we compare performance of the sparse latent representations with their dense counterpart on the  text classification tasks (\S\ref{sec:classification}). Second, the stability of sparsification of HSVAE is compared with the state-of-the-art MAT-VAE (\S\ref{sec:rsparse}). Then, to better understand performance of our model on the downstream task, we examine the sparsity patterns (\S\ref{sec:analysis}).
\paragraph{Remark.} An integral part of the experiments is the analysis of the learned representations.  In this sense, tasks that rely on understanding of semantics (e.g., GLUE \cite{wang-etal-2018-glue}) or syntax (e.g., \cite{marvin-linzen-2018-targeted}) would be non-trivial to analyse due to their inherent complexity. We consider classification tasks because the distribution of words alone could  be a good indicator of class labels. Given the unsupervised nature of the models, we explore if this surface-level distribution of words could be captured by the sparsity patterns in the learned representation.


\subsection{Experimental Setup}
\subsubsection{Corpora Preprocessing}
 We use Yelp\footnote{\url{https://github.com/jxhe/vae-lagging-encoder/blob/master/prepare_data.py}.} as it is, without any additional preprocessing. As for DBpedia\footnote{\url{https://github.com/srhrshr/torchDatasets/blob/master/dbpedia_csv.tar.gz}} and Yahoo\footnote{\url{https://github.com/jxhe/vae-lagging-encoder/blob/master/prepare_data.py}.}, the preprocessing is as follows: (1) removing all non-ASCII characters, quotations marks, and hyperlinks, (2) tokenising with spaCy\footnote{\url{https://spacy.io}}, (3) lower-case conversion for all tokens, then (4) \textbf{for each class} we randomly sample 10,000 sentences for the training corpus and 1,000 sentences for the test and validation respectively. The vocabulary  size of the both corpora is reduced to the first 20,000 most frequent words.

\subsubsection{Baselines and Models} To ground the performance of HSVAE  we use 4 baselines: 1) VAE is a version of the vanilla VAE used in \citet{higgins2016},  2) the same VAE model but the activation of $\mu$ and $\sigma$ of $q_{\phi}(z|x)$ regularised by either $L^1$ (VAE\textsubscript{$L^1$}) or $L^2$ (VAE\textsubscript{$L^2$}) norms,  3) MAT-VAE is a VAE framework introduced by \citet{pmlr-v97-mathieu19a} and 4) simple classifier which is simply a text encoder with a classifier on top of it. For all these models we use a GRU network \cite{DBLP:journals/corr/ChoMGBSB14} to encode and decode text sequences. We set the dimesnionality of the both encoder and the decoder GRU's to  512D and the dimensionality of the word embeddings is 256D. The decoder and the encoder share the word embeddings.  To train the model we use the Adam optimiser \cite{adam} with the learning rate: 0.0008. 

\paragraph{BERT vs GRU Encoder.} Inspired by \citet{li2020optimus}, we replace the GRU network used in VAE and HSVAE encoders with a pretrained BERT\footnote{After extracting features from a sequence with BERT, we then applying MLPs to extract features for the posterior distributions, as it is the case for the encoder with GRU network. } \cite{devlin-etal-2019-bert}, while keeping the GRU decoder. We refer to these models as  B-VAE and B-HSVAE, respectively. Also, we compare the task performance of these VAE models with the plain pretrained base-BERT\footnote{\url{https://huggingface.co/transformers/model_doc/bert.html}}. To train B-VAE and B-HSVAE, we use the Adam optimiser with the learning rate: 0.00008.

\paragraph{Dimensionality of $z$.} 
We use the following two dimensions: 32D and 768D. Since, HSVAE and MAT-VAE induce sparse latent representations we want to make sure that they perform robustly regardless of the number of the dimensions.  

\paragraph{KL-Collapse.} None of the used VAE models is immune to the KL-collapse \cite{DBLP:journals/corr/BowmanVVDJB15} - when the KL term becomes zero and the decoder ignores the information provided by the encoder through $z$. To address this issue, in all the models, we put a scalar value $\psi, \lambda < 1$ on the KL terms of the VAE's objective function \cite{he2018lagging}.

\paragraph{Coupling Encoder with Decoder.} To connect the encoder with the decoder we concatenate the latent variable $z$, sampled from the posterior distribution, to word embeddings of the decoder at each time step \cite{prokhorov-etal-2019-importance}. Also,  for GRU encoders we take the last hidden state to parameterise the posterior distribution. For BERT encoder, we take average pooling of all token's embeddings produced by the last layer of BERT.

\subsubsection{Evaluation Metrics}
\paragraph{Text Classification.} To report the classification performance we use accuracy as a metric.
\paragraph{Sparsity.} We measure Hoyer \cite{Hoyer_metric} on the representations of all data points in a corpus and report its average as our sparsity metric \cite{pmlr-v97-mathieu19a}. Hoyer, in a nutshell, is ratio of the $L^2$ to $L^1$ norm, normalised by the number of dimensions. Higher indicates more sparsity. More specifically, to evaluate the average Hoyer, or as we refer to it as Average Hoyer (AH) in the experiments, either on a validation or test corpus we employ the following procedure. First, for each $x_i$ in the corpus $\{ x_1,...,x_n \}$  we  obtain its corresponding $z_i$ by sampling it form a probabilistic encoder of a VAE model, such that for each $x_i$ we sample one $z_i$: e.g. $x_1 \xrightarrow{\makebox[0.01cm]{}} z_1$. Then we normalise $\bar{z}_i = z_i / \sigma(z)$, where $z = \{ z_1,...,z_n \}$, and $\sigma(.)$ is the standard deviation. Finally, for each $\bar{z}_i$ we compute  Hoyer as follows:

\begin{equation}
 Hoyer(\bar{z}_i)=\frac{\sqrt{d} - ||\bar{z}_i||_1/||\bar{z}_i||_2 }{\sqrt{d} -1},   
\end{equation}
where $d$ is the dimensionality of $\bar{z}_i$. To report the Hoyer for the whole corpus we compute the Average Hoyer $=\frac{1}{N}\sum_i^N Hoyer(\bar{z}_i)$, where $N$ is a number data points in a test or validation corpus.

\begin{figure}[t!]%
\centering
\includegraphics[scale=0.9]{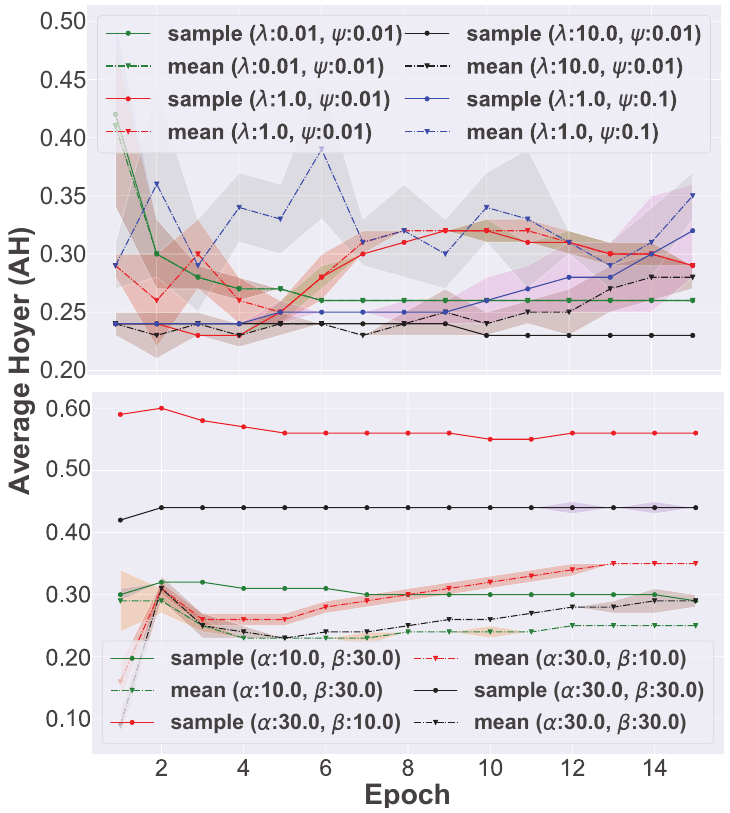}
\vspace*{-2.3mm}
\caption{Average Hoyer (AH) on DBpedia corpus dev set  for different parameterisations of \citet{pmlr-v97-mathieu19a} (Top) vs. HSVAE (Bottom). Same is  observed on Yelp and Yahoo (see Appendix \ref{appendix:hoyer}). Lines are an average over the 3 runs of the models, the shaded area is the standard deviation. The dimensionality of the latent variable of the models is 32D.}
\label{figure:hoyer}
\end{figure}

\subsection{Text Classification}\label{sec:classification}
Prior to use of a VAE encoder in the classification experiment, we pretrained it using the full VAE model with the corresponding VAE's objective function on one of the target corpus: Yelp, Yahoo or DBpedia. We compare performance of the sparse latent representations with their dense counterparts  on the three text classification tasks (Figure \ref{figure:acc_hoyer}). The classifier that we use comprises of the two dense layer of 32D each with the Leaky ReLU \cite{Maas2013RectifierNI} activation function. To establish whether the performance gain or loss on the tasks is achieved thanks to the sparsity inductive bias,  for all the VAE models and BERT we freeze the parameters of the encoder and only train the classifier which we put on top of the encoder. However, for the simple classifier model its text encoder is being trained together with the classifier. When the classifier, $p(y|x)$, is trained with a probabilistic VAE encoder we marginalise the latent variable(s). This is done for instance for HSVAE as, $$p(y|x) = \int_{z,\gamma} ~p(y|z)q(z|x,\gamma)q_\phi(\gamma|x)dzd\gamma$$ 

We approximate the integral with MC by taking $K=5$ samples from the probabilistic encoder both to train and to test the classifier: For each $x_i$ in a batch $\{ x_1,...,x_p \}$:
\begin{enumerate}
    \item sample $K$ of  $\gamma_{i,j}$ from $q_\phi(\gamma|x_i)$ i.e. a set of sampled $\gamma$'s is $\{\gamma_{i,1},...,\gamma_{i,K} \}$ 
    \item sample $K$ of $z_{i,j}$ from $q_\phi(z|x_i, \gamma_{i,j})$ i.e. a set of sampled tuples of $z_{i,j}$ and  $\gamma_{i,j}$ is $\{(z_{i,1}, \gamma_{i,1}),...,(z_{i,K}, \gamma_{i,K}) \}$ in other words for each $\gamma_{i,j}$ we sample only one $z_{i,j}$.
\end{enumerate}
For the other VAEs the procedure is similar. With the MC approximation : $p(y|x) \approx 0.2\times\sum_{i}^5 ~p(y|z_i)$.

For a systematic comparison of various VAEs, we collate classification performance of VAEs with comparable reconstruction loss  - which indicates how informative the latent code is for the decoder during reconstruction. In other words the reconstruction loss serves as an intrinsic metric. Thus, for an example, in Figure \ref{fig:Ng1}, for the Yelp corpus all the VAE models have a similar reconstruction loss. The same applies to Figure \ref{fig:Ng2} and Figure \ref{fig:Ng3}.

Comparing the accuracy of the classifiers that are trained with the different latent representations i.e. sparse and dense  (Figure\ref{figure:acc_hoyer}), shows that in general the performance of the sparse latent representations induced by HSVAE or MAT-VAE is on par with their dense latent counterparts inferred by the VAEs. However, the performance of HSVAE slightly lagging behind on the Yelp corpus when the dimensionality  of the latent representation is 32D (Figure \ref{fig:Ng1}).  We put forward a hypothesis that may explain this in  Section \ref{sec:analysis}.  Also, when the dimensionality of the latent representation is 32D, the accuracy of MAT-VAE is slightly better than of HSVAE, but this performance is reached at lower levels of sparsity. Additionally, we found that regularising  the posterior parameters of the VAE model with either $L^1$ or $L^2$ norm, in some cases, helps to increase the classification accuracy, but does not reach AH higher than the vanilla VAE. Notably, the classification performance of all the VAE models becomes almost identical when the dimensionality of the latent space is increased from 32D to 768D, with HSVAE slightly outperforming all other VAEs on the DBpedia corpus (Figure \ref{fig:Ng2}). We further elaborate on it in Section \ref{sec:analysis}. 
 
 Use of BERT as an encoder, in our settings, only gives an improvement on the Yahoo corpus with B-HSVAE performing on par with B-VAE, but does not reach the classification accuracy of the plain BERT. We hypothesise that to reach the full potential of the use of a pretrained encoder in a VAE model one needs to pair it with a powerful decoder such as GPT-2 \cite{radford2019language} as it is the case in the \citet{li2020optimus} VAE model. Further exploration of this was beyond our compute resource.

Finally,  one can observe that the simple classifier model performs on a par (in Figure \ref{fig:Ng1}) or even worse (Figure \ref{fig:Ng2} ) than the VAE models on the Yelp corpus. Putting it into the context that the VAE encoders are not being trained with a supervision signal while the encoder of the simple classifier is, we speculate that this can be explained by the discussion put forward in \citet{Valpola2014FromNP}. A classifier in nature tries to remove all the information that is not relevant to the supervision signal, while an autoencoder tries to preserve as much as possible information in the latent code in order to reconstruct the original input data reliably. Thus, if the  distribution of class related words in a text alone (see \cref{sec:class_kl_div}) is not indicative enough of a class then the classifier may perform poorly. In our case, we hypothesise that the VAE models capture some additional information other than class distribution of words in text that allows it to better discriminate the classes.  For example, some class may have shorter sentences, on average, than the sentences presented in the other classes. This may provide an additional bias that allows the VAE models to discriminate sentences from this class from the sentences from the other classes. Thus, with this additional bias VAEs can perform better than the simple classifier.  We leave this investigation for a future work. 


\subsection{Representation Sparsity}\label{sec:rsparse}
In Figure~\ref{figure:hoyer} we compare HSVAE with MAT-VAE. We report AH both on the mean
and samples from the posterior distributions. As illustrated, MAT-VAE struggles to achieve steady and consistent AH regardless of the configurations of its hyperparameters ($\psi,\lambda$). However,  HSVAE stably controls the level of sparsity with $\alpha$ and $\beta$ parameters, a positive effect of its more flexible posterior distribution and the learnable distribution over $\gamma$.

\begin{figure}[t!]
\centering
\begin{subfigure}[b]{0.48\textwidth}
\centering
   \includegraphics[
height=3.0in, width=2.6in, 
angle=270]{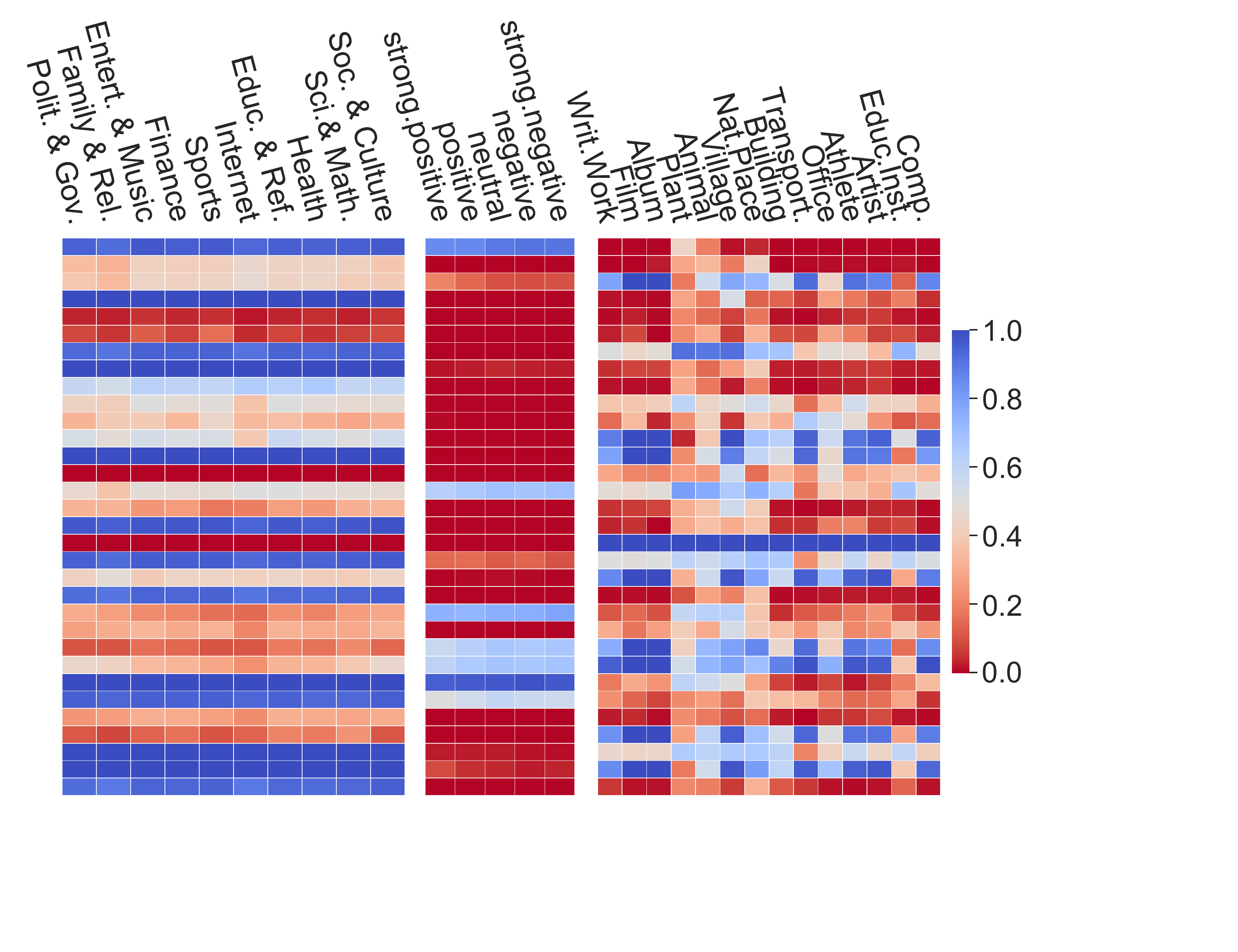}
   \caption{}
   \label{fig:gamma_1} 
\end{subfigure}

\begin{subfigure}[b]{0.48\textwidth}
\centering
  \includegraphics[
height=2.5in, width=2.6in, 
angle=270]{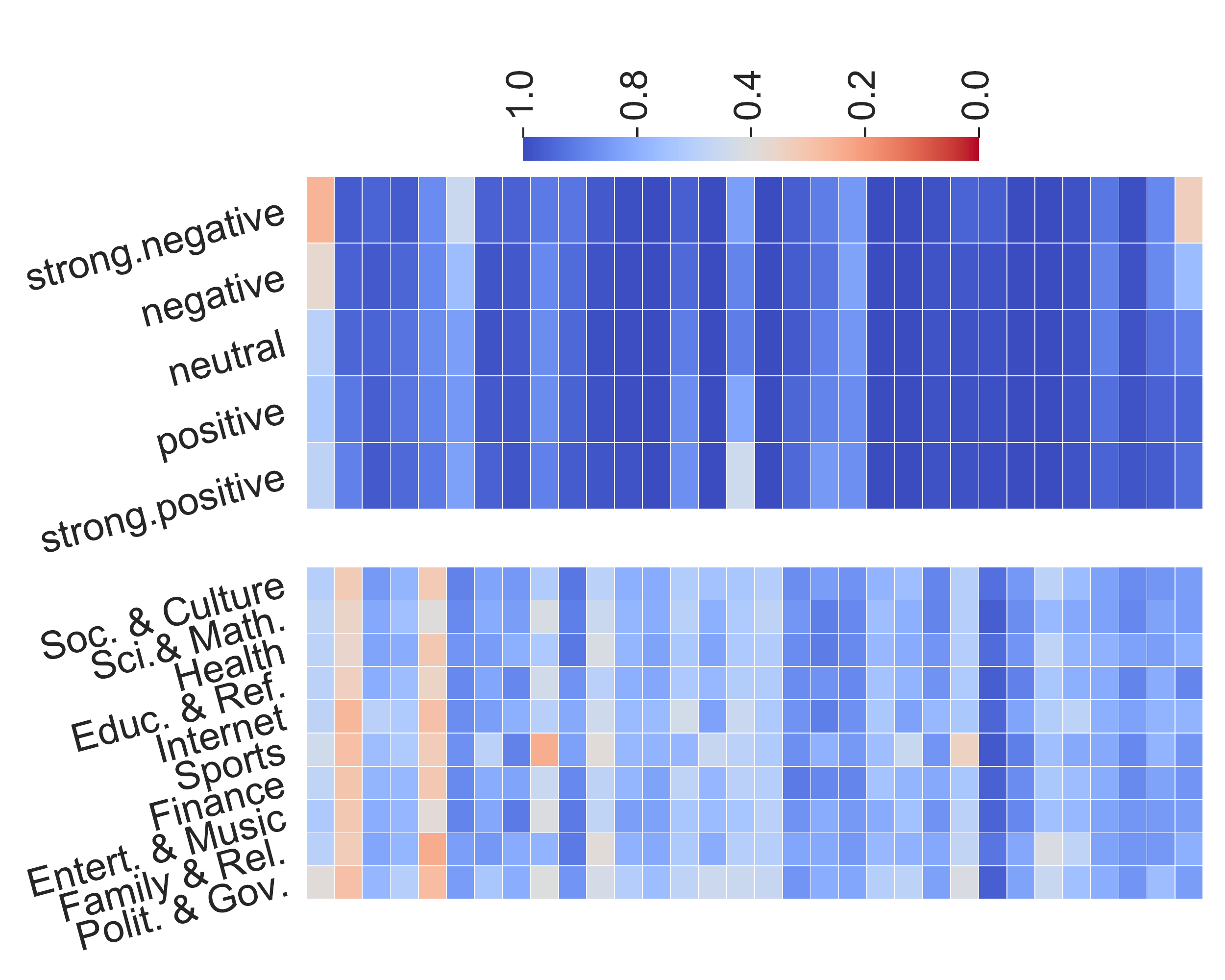}
   \caption{}
   \label{fig:gamma_2}
\end{subfigure}
\vspace{-7mm}
\caption[]{Heat maps of $\gamma_{class}$ (Section \cref{sec:analysis}). (a) $\gamma_{class}$ of 32D  - from left to right: Yahoo, Yelp, DBpedia. (b) contiguous 32D out of 768D of $\gamma_{class}$  - from left to right: Yahoo, Yelp. }
\label{figure:class_distribution}
\end{figure}

\subsection{Can Sparsity Patterns Encode Classes?}\label{sec:analysis}
In order to identify pertinent features, the unsupervised representation learning models are typically trained/fine-tuned on corpora that are closely related to the downstream task. As such, without a supervisory signal, the model can only rely on the distribution of words in a text in order to identify these relevant features for the task.  Ideally, compared to their dense counterparts, an unsupervised sparsification model such as HSVAE could result in performance improvement on downstream tasks if they capture the task-related  features and discard the noisy features. However, if the sparsification model fail to capture the task related signal in its sparsity pattern; it can hurt the performance of the model on the downstream task as the task-related information can be removed. In what follows we investigate this direction by analysing the sparsity patterns and  relate this analysis to the classification performance of the model (\cref{sec:classification}).

\begin{figure*}[ht!]
\hspace*{-3mm}
   \subfloat[\label{fig:1}]{%
      \includegraphics[trim=20 40 50 30,clip, width=0.39\textwidth]{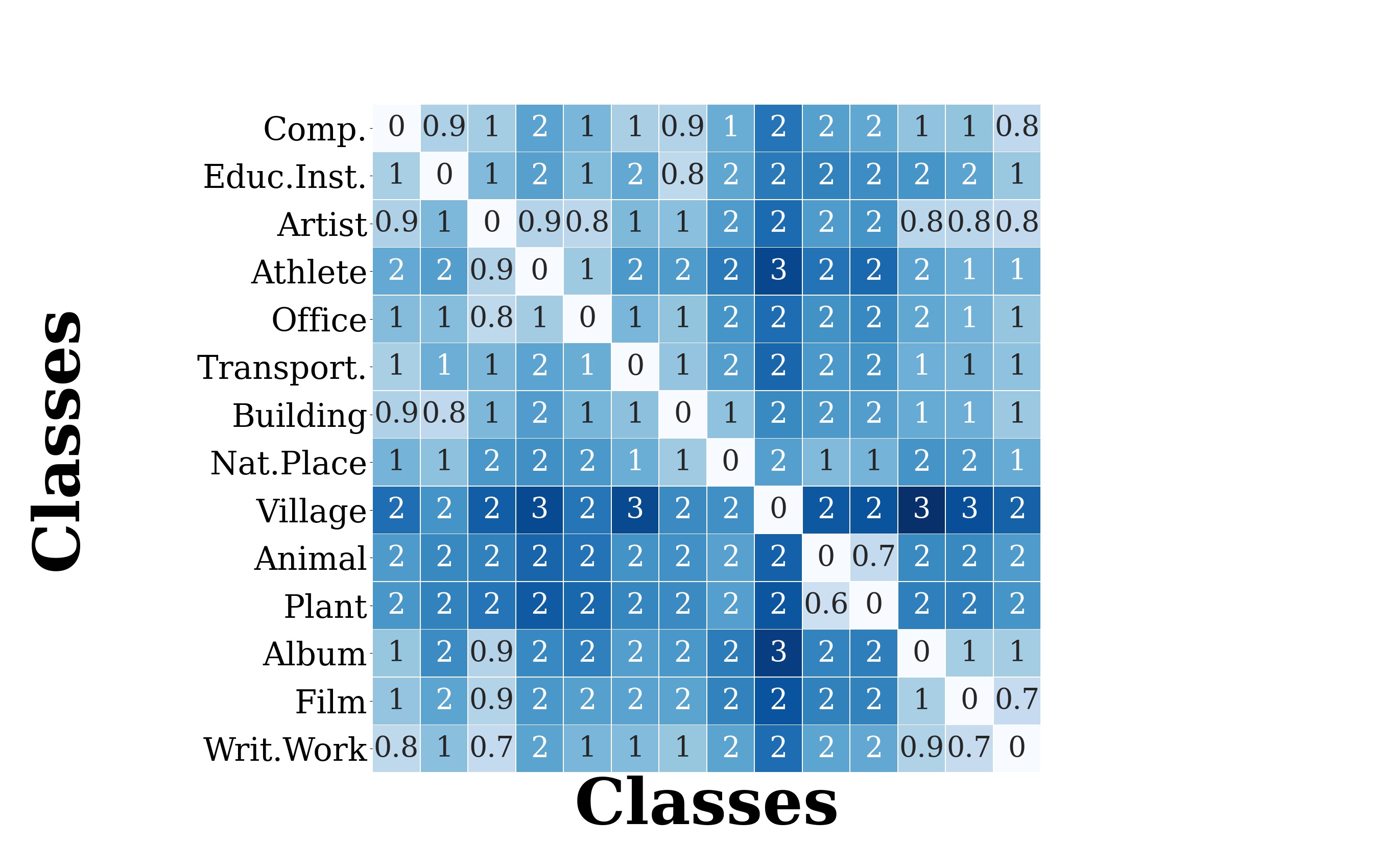}}
\hspace*{-14mm}
   \subfloat[\label{fig:2} ]{%
      \includegraphics[trim=10 40 10 30,clip, width=0.39\textwidth]{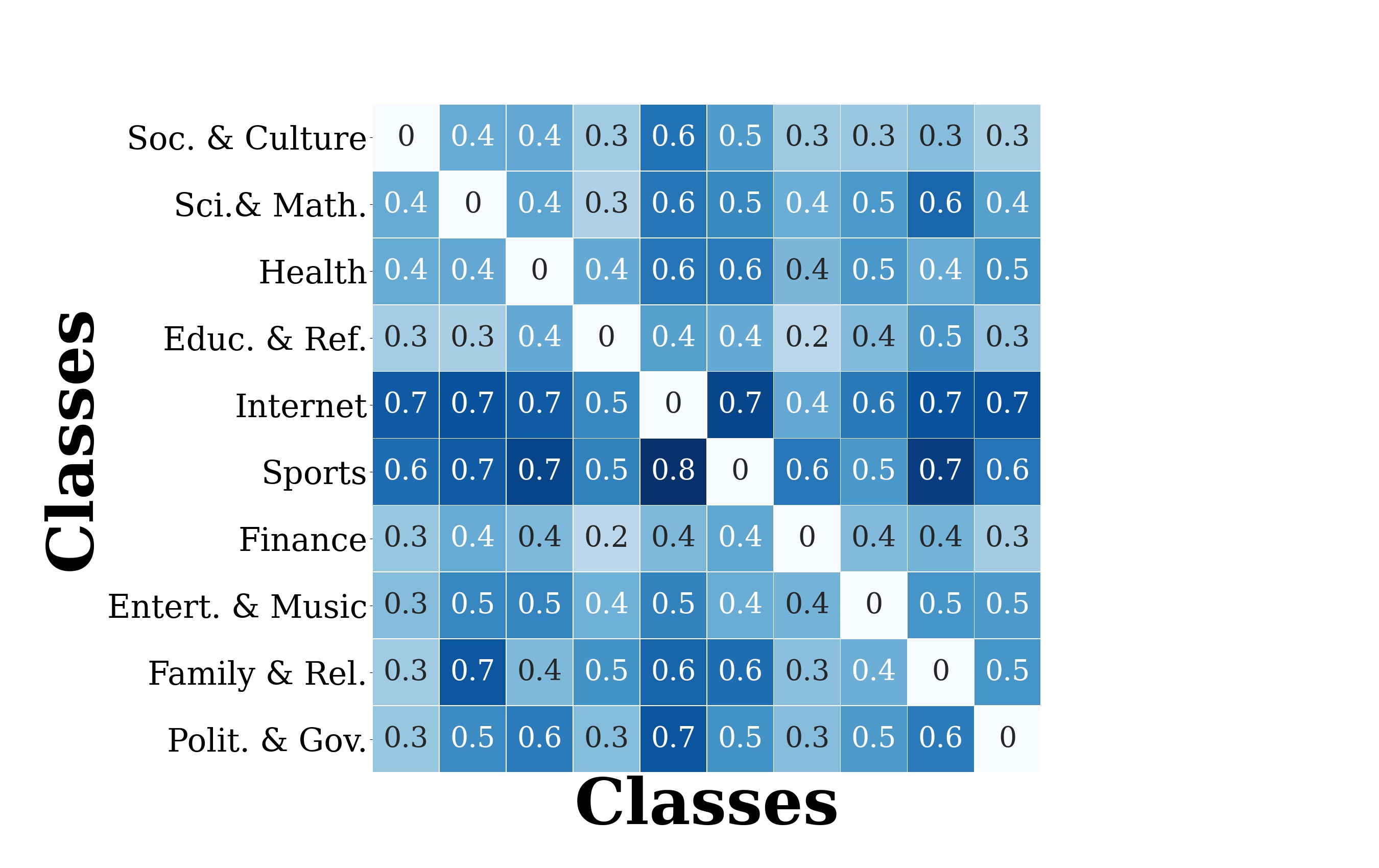}}
\hspace*{-15mm}
   \subfloat[\label{fig:3}]{%
      \includegraphics[trim=30 40 30 30,clip, width=0.39\textwidth]{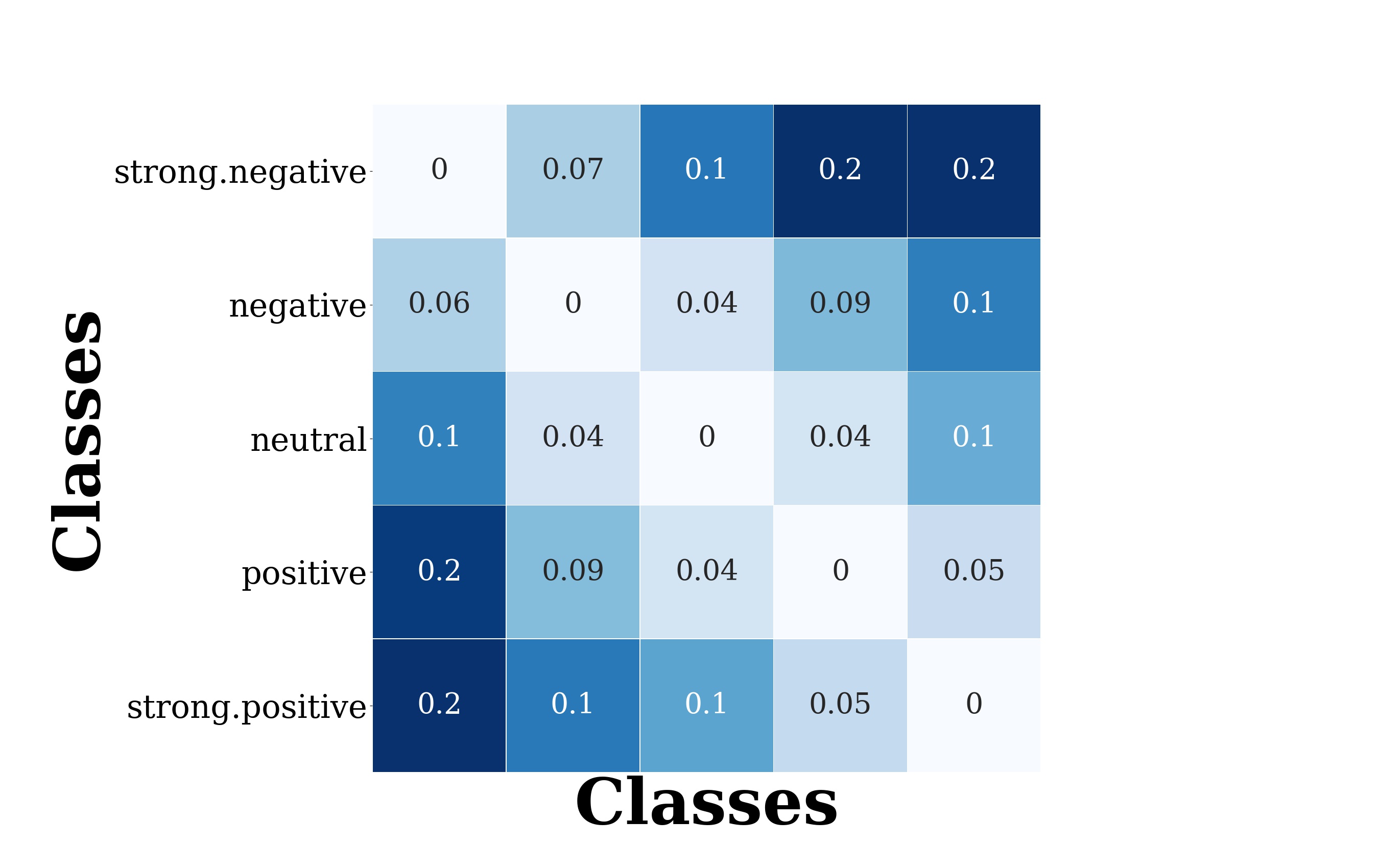}}\\
\vspace{-3mm}
\caption{Experimental results for KL between classes on the three corpora: DBpedia (a), Yahoo (b) and Yelp (c).}
\label{fig:was_unimodal_random_gamma}
\end{figure*}
\paragraph{Analysis of $\gamma$.} We hypothesise that if $\gamma$ captures a class of a sentence then the sentences that belong to the same class should have a similar sparsity patterns in $\gamma$. To obtain a class specific $\gamma_{class}$,  first, for each sentence $x$ we obtain the mean of the posterior distribution: $q_\phi(\gamma|x)$ and we denote it as  $\mu_{\gamma(x)}$. Then we binarise the mean such as $\mu_{\gamma(x)}^b$ = Binarise($\mu_{\gamma(x)}$), where Binarise($\cdot$) is defined as: 0 if $\mu_{\gamma(x)}$ $<$ 0.5 and 1 otherwise.   Finally, for each class we average its $\mu_{\gamma(x)}^b$ vectors to obtain a single vector that represent this class: $\gamma_{class} = \frac{1}{M}\sum_{x \in class} \mu_{\gamma(x)}^b,$
where $M$ is a number of sentences in the class. The averaging removes the information that differentiate these sentences, while preserving the class information that is shared among them. A similar approach was also used in \citet{pmlr-v97-mathieu19a}.

Figure \ref{figure:class_distribution}  reports the magnitudes of the $\gamma_{class}$ vectors as heat maps for the three corpora. One would expect that  $\gamma_{class}$ of  different classes should differ. For 32D $\gamma_{class}$ (Figure \ref{fig:gamma_1}) this is the case when HSVAE is trained on the DBpedia and Yahoo but not on Yelp.  Taking into account the unsupervised nature of these models,  this difference is echoing the distribution of words in the classes, which is more distinct in DBpedia and Yahoo, but not in Yelp (see \cref{sec:class_kl_div}). We also hypothesis that this observation can explain inferior performance of the model on the Yelp corpus (Figure \ref{fig:Ng1}).

In contrast, for $\gamma_{class}$ in 768D (Figure \ref{fig:gamma_2}) one can observe that the different classes have different activation patterns even when  HSVAE is trained on the Yelp corpus.\footnote{ In Figure \ref{fig:gamma_2} we only show 32D out of 768D. This is one of the subsets of the 768 dimensions where the distributedness is present. It is not unique and the distributedness is also present in other dimensions of the 768D code.}  Also, the distributedness of the activation patterns now becomes more apparent when HSVAE is trained on the Yahoo corpus. This observation is also related to the distribution of words in the text (further elaborated in \cref{sec:class_kl_div}). 

Intuitively, to reconstruct a sentence a VAE model first captures aspect of data that are the most conducive for reconstruction error reduction~\cite{DBLP:journals/corr/abs-1804-03599}. 
Therefore, given the limited dimensionality of the latent vector, the model will prioritised aspects of data during encoding. As such, if the information such as sentence class is not strongly presented in the corpus the model could potentially ignore it during encoding. However, when the dimensionality of the latent space is increased, the model has more capacity to represent various aspects of data that may otherwise be ignore in the smaller dimensionality. We speculate this could explain the presence of distributedness of $\gamma_{class}$ on Yelp for 768D as opposed to 32D, which also translates into matching the task performance of its dense counterpart (Figure \ref{fig:Ng2}).

\subsubsection{Class Kullback–Leibler Divergence}
\label{sec:class_kl_div}
The question that has yet not been addressed is why in some cases the HSVAE model is more successful at capturing the class distribution when trained on DBpedia compared to Yelp.  We previously hypothesised that the reason for this can be a word distribution in a text. To empirically test our hypothesis, we calculate the add-1 smoothed probabilities of words in the classes and measure the pairwise KL divergence across them. The magnitudes of the pairwise KL divergences are shown in Figure \ref{fig:was_unimodal_random_gamma}. As demonstrated, the magnitude of the KL divergence is the largest for DBpedia and smallest for Yelp. This indicates that separating classes in Yelp would rely on more subtle aspects of data, whereas surface-level cues are more present in DBpedia and allow for an easier discrimination. 

\vspace{-1.1mm}
\section{Related Work}
\label{sec:related work}
Learning sparse representations of data can be dated back to  \citet{Olshausen_1996}. This work motivates encoding of images in sparse linear codes for its biological plausibility and efficiency. It was later argued by \citet{10.1561/2200000006} that compared to the dimensionality reduction approaches, sparsity is a more efficient method for representation learning on vectors with fixed dimensionality.

\paragraph{Representation Sparsity.} In NLP, learning sparse representations has been explored for various units of text with most of the focus placed on sparse representation of words. 
As the earliest work that moved in this direction, \citet{murphy-etal-2012-learning}
looked into sparse representations for ease of analysis, performance, and being more cognitively plausible. This idea was further developed by many other researchers ~\cite{faruqui-dyer-2015-non, DBLP:conf/icml/YogatamaFDS15, faruqui-etal-2015-sparse,DBLP:conf/ijcai/SunGLXC16, DBLP:conf/aaai/SubramanianPJBH18, arora-etal-2018-linear, DBLP:journals/corr/abs-1911-01625}. Sparsification of the large units of text (i.e., sentences) has not received a lot of attention, perhaps due to inherent complexity of sentence/phrase representations: i.e., encoding and analysing syntactic and semantic information in a sentence embedding is rather a non-trivial task. To the best of our knowledge, the only model that sparsifies sentence emebeddings is introduced by \citet{DBLP:conf/emnlp/TrifonovGPH18}. The authors introduced a Seq2Seq model \cite{10.5555/2969033.2969173} with the Sparsemax  layer \cite{pmlr-v48-martins16} between the encoder and the decoder which induces sparse latent codes of text. This layer allows to learn codes that can be easier to analyse compared to their dense counterparts, but  it is limited to modelling the categorical distribution. Thus restricts a type a sentence representations that can be learned. 

\paragraph{VAE-based Representation Sparsity.} 
VAE-based sentence representation learning has shown superior properties compared to their deterministic counterparts on tasks such as text generation~\cite{DBLP:journals/corr/BowmanVVDJB15}, Semantic Textual Similarity~\cite{li-etal-2020-sentence} and other wide range of language tasks \cite{li2020optimus}. 
While a handful of VAE-based sparsification methods have been proposed recently~\citet{pmlr-v97-mathieu19a}~(\textsc{mat}), \citet{tonolini2019vsc}~(\textsc{ton}), they have been only evaluated on image domain. We summarise the similarity and key differences with HSVAE model: 
\begin{description}
\item [\textsc{Prior and Posterior.}] All three frameworks use the Spike-and-Slab distribution to construct the prior on $z$. While the posterior distribution in \textsc{mat} remains as a Gaussian, both \textsc{ton} and HSVAE opt for Spike-and-Slab. 
However, \textsc{ton} controls the sparsity level in an indirect way via ``pseudo data''~\cite{Tomczak2018VAEWA} used in prior, whereas HSVAE's probabilistic treatment of $\gamma$ enables direct control on the target sparsity level.
\item [\textsc{Objective.}] HSVAE is trained with a principled ELBO (eq.~\ref{hsvae-objective}), while the other two add additional regularisers to the ELBO of VAE (eq.~\ref{elboobjective}). For instance, \textsc{mat} add a \emph{maximum mean discrepancy} (MMD) divergence between $z$'s aggregated posterior and prior $\text{MMD}(q_\phi(z),p_\theta(z))$ and include scalar $\psi$ and $\lambda$ weights to the KL and $\text{MMD}$ term, respectively, see Appendix \ref{appendix:obj_funct}.

\end{description}

\paragraph{Model Sparsity.} Concurrent to the widespread use of large models such as Transformers~\cite{46201} in NLP, sparsification of these models is also becoming popular~\cite{zhang2020sparsifying, zhao2019explicit, correia-etal-2019-adaptively, ye2019bptransformer, child2019generating}. The most common approach to sparsify a Transformer is to reduce a number of connection between the words/tokens in the self attention kernel e.g. \citet{correia-etal-2019-adaptively}.  However, these approaches still learn dense continuous representations of token/word/sentence embeddings.

\vspace{-1.1mm}
\section{Conclusion}
We provided a sober analysis of several unsupervised sparsification frameworks based on VAEs, both in terms of the impact on downstream tasks and the level of sparsity achieved.    Also, we presented a novel VAE model - Hierarchical Sparse Variational Autoencoder (HSVAE), outperforming existing SOTA model~\citep{pmlr-v97-mathieu19a}. 
Ideally, sparse representations should be capable of  encoding the underlying characteristics of a corpus (e.g. class), in activation patterns as shown to be the case for HSVAE. Moreover, using the text classification corpora as a testbed, we established how statistical properties of a corpus such as word distribution in a class affect the ability of learned sparse codes to represent task-related information.

Moving forward, HSVAE model along with the analysis provided in this paper can serve as a good basis for the design of sparse models that induce continuous sparse vectors of text. For example, a potential extension of HSVAE could be  an incorporation of explicit linguistic biases into the learned representations with the group sparsity \cite{DBLP:conf/icml/YogatamaFDS15}. Furthermore, as we discussed in Section \ref{sec:related work}, sparsity found its application in the Transformers, but it, mainly, has been used to reduce the number of connection between the words/tokens. With the HSVAE framework one can also learn sparse continuous representations of token/word/sentence embeddings.

\bibliographystyle{acl_natbib}
\bibliography{acl2021}

\appendix

\section{Derivations of ELBO}
\label{appendix:elbo_derivation}
Starting from the $\mathbb{D}_{KL}(q_\phi(z,\gamma|x)||p_\theta(z,\gamma|x))$, we derive the Evidence Lower Bound (ELBO) as follows:
\begin{equation}
\label{eq_1:KL}
\resizebox{.65\linewidth}{!}{%
$\begin{aligned}
 \mathbb{D}_{KL}(q_{\phi}(z,\gamma|x)||p_{\theta}(z,\gamma|x)) = \\ \int\limits_{z,\gamma} dz d\gamma\, q_{\phi}(z,\gamma|x)\log\frac{q_{\phi}(z,\gamma|x)}{p_{\theta}(z,\gamma|x)},
 \end{aligned}$%
 }
\end{equation}
after rearranging terms in equation \ref{eq_1:KL} we can obtain:
\begin{equation}
\resizebox{.80\linewidth}{!}{%
$\begin{aligned}
  \log p_{\theta}(x) -  \mathbb{D}_{KL}(q_{\phi}(z,\gamma|x)||p_{\theta}(z,\gamma|x)) = \\ \underbrace{\int\limits_{z,\gamma} dz d\gamma\, q_{\phi}(z,\gamma|x)\log\frac{p_{\theta}(z,\gamma, x)}{q_{\phi}(z,\gamma|x)}}_{\text{ELBO}},
  \end{aligned}$%
  }
  \label{eq_2:ELBO}
\end{equation}{}

Based on the independence assumption that we make in our graphical model (Figure 1) the generative model factorises as: $p_{\theta}(z,\gamma,x)=p_{\theta}(x|z)p_{\theta}(z|\gamma)p_{\theta}(\gamma)$ and the inference model factorises as: $q_{\phi}(z,\gamma|x)=q_{\phi}(z|\gamma,x)q_{\phi}(\gamma|x)$. Therefore, we can rewrite the ELBO as follows:
\begin{equation}
\resizebox{.96\linewidth}{!}{%
$ \int\limits_{z,\gamma} dz d\gamma\, q_{\phi}(z|\gamma,x)q_{\phi}(\gamma|x)\log\frac{p_{\theta}(x|z)p_{\theta}(z|\gamma)p_{\theta}(\gamma)}{q_{\phi}(z|\gamma,x)q_{\phi}(\gamma|x)},
 $%
 }
  \label{eq_3:ELBO}
\end{equation}{}
We can further rewrite the ELBO as a sum of the three separate terms. Where the first term is:
\begin{equation}
\resizebox{.83\linewidth}{!}{%
$\begin{aligned}{}
&\int\limits_{z,\gamma}  dz d\gamma\, q_{\phi}(z|x, \gamma) q_{\phi}(\gamma|x)\log p_{\theta}(x|z) \\
&\int\limits_{\gamma} d\gamma\, q_{\phi}(\gamma|x) \int\limits_{z} dz \,  q_{\phi}(z|x, \gamma) \log p_{\theta}(x|z) \therefore \\
&\bigg \langle  \int\limits_{z} dz \, q_{\phi}(z|x, \gamma) \log p_{\theta}(x|z) \bigg \rangle_{q_{\phi}(\gamma|x)} \therefore
\end{aligned}$%
}
\end{equation}
The second term is:
\begin{equation}
\resizebox{.98\linewidth}{!}{%
$\begin{aligned}{}
&\int\limits_{z,\gamma} dz d\gamma\, q_{\phi}(z|x, \gamma) q_{\phi}(\gamma|x)[\log q_{\phi}(z|x, \gamma) - \log p_{\theta}(z|\gamma)] \\
&\bigg \langle \int\limits_{z} dz \, q_{\phi}(z|x, \gamma)[\log q_{\phi}(z|x, \gamma) - \log p_{\theta}(z|\gamma)] \bigg \rangle_{q_{\phi}(\gamma|x)}  \therefore \\
&\bigg \langle \mathbb{D}_{KL}(q_{\phi}(z|x,\gamma)||p_{\theta}(z|\gamma)) \bigg \rangle_{q_{\phi}(\gamma|x)}  \therefore
\end{aligned}$%
}
\end{equation}{}
Finally, the third term is:
\begin{equation}
\resizebox{.98\linewidth}{!}{%
$\begin{aligned}{}
&\int\limits_{z,\gamma} dz d\gamma\, q_{\phi}(z|x, \gamma) q_{\phi}(\gamma|x)[\log q_{\phi}(\gamma|x) - \log p_{\theta}(\gamma)] \\ 
&\int\limits_{\gamma} d\gamma\, q_{\phi}(\gamma|x)[\log q_{\theta}(\gamma|x) - \log p_{\theta}(\gamma)] \underbrace{\int\limits_{z}dz \, q_{\phi}(z|x, \gamma)}_{\text{sums to 1 for each}: \gamma}  \therefore \\ 
&\int\limits_{\gamma} d\gamma\, q_{\phi}(\gamma|x)[\log q_{\phi}(\gamma|x) - \log p_{\theta}(\gamma)]  \therefore \\
&\mathbb{D}_{KL}(q_{\phi}(\gamma|x)||p_{\theta}(\gamma))  \therefore
\end{aligned}$%
}
\end{equation}{}
Collecting all the three terms into the single ELBO:
\begin{equation}
\resizebox{.74\linewidth}{!}{%
$\begin{aligned}
 \bigg \langle  \int\limits_{z} dz \, q_{\phi}(z|x, \gamma) \log p_{\theta}(x|z) \bigg \rangle_{q_{\phi}(\gamma|x)}  -\\ -\bigg \langle \mathbb{D}_{KL}(q_{\phi}(z|x, \gamma)||p_{\theta}(z|\gamma)) \bigg \rangle_{q_{\phi}(\gamma|x)}-\\ -\mathbb{D}_{KL}(q_{\phi}(\gamma|x)||p_{\theta}(\gamma)),
\end{aligned}$%
}
\end{equation}

\section{Objective Functions of \citet{pmlr-v97-mathieu19a}  and \citet{tonolini2019vsc} Models}
\label{appendix:obj_funct}
The objective function of \citet{pmlr-v97-mathieu19a} is:
\begin{equation*}
\resizebox{.9\linewidth}{!}{%
$\begin{aligned}
\big \langle \log p_\theta({x}|{z}) \big \rangle_{q_\phi({z}|{x})}  -
\psi KL(q_{\phi}(z|x)||p_\theta(z)) -\\- \lambda \mathbb{D}(q_{\phi}(z),p_\theta(z)),
\end{aligned}$%
}
\end{equation*}
where $\psi$ and $\lambda$ are the scalar weight on the terms and \citet{tonolini2019vsc} is:
\begin{equation*}
\resizebox{.9\linewidth}{!}{%
$\begin{aligned}
\big \langle \log p_\theta({x}|{z}) \big \rangle_{q_\phi({z}|{x})}  -
 KL(q_{\phi}(z|x)||q_\phi({z}|{x_u}) -\\- J\times\mathbb{D}_{KL}\big(\Bar{\gamma}_u || \alpha)\big),
\end{aligned}$%
}
\end{equation*}
where $J$ is the dimensionality of the latent variable $z$, $x_u$ is a learnable pseudo-input \cite{Tomczak2018VAEWA} and $\alpha$ is prior sparsity.

\section{Deriving Marginal of (Univariate) Spike-and-Slab Prior}
 We derive the Spike-and-Slab distribution  by integrating out the index component which is distributed  as a Bernoulli variable. This result is quite well-known in machine learning, however for the ease of the reader we present it here as a quick reference.

The derivation: assume 1) $\pi \sim p(\pi;\gamma)$ is a $Bernoulli(\gamma)$ and 2)
$p(z|\pi) =  (1-\pi) \times p_1(z) + \pi \times p_2(z)$, where  $p_1(z) \sim \mathcal{N}(z; 0, 1)$ and $p_2(z) \sim \mathcal{N}(z; 0,\sigma \rightarrow 0)$  is a Spike-and-Slab model. The the marginal Spike-and-Slab prior over $z$ can be obtained in the following way:
\begin{equation*}
\resizebox{.98\linewidth}{!}{%
$\begin{aligned}
   & p(z;\gamma) = \sum_{i=0}^1p(z|\pi=i) p(\pi=i;\gamma)\\
    & p(z|\pi=0) p(\pi=0;\gamma) + p(z|\pi=1) p(\pi=1;\gamma) \therefore\\
     &[(1-0) \times p_1(z) + 0 \times p_2(z)] p(\pi=0;\gamma) +\\&+ [(1-1) \times p_1(z) + 1 \times p_2(z)] p(\pi=1;\gamma) \therefore
\end{aligned}$%
}
\end{equation*}
Expanding brackets:
\begin{equation*}
\resizebox{.98\linewidth}{!}{%
$\begin{aligned}
    & p_1(z)p(\pi=0;\gamma) + p_2(z)p(\pi=1;\gamma)\therefore\\
    & \mathcal{N}(z; 0, 1)p(\pi=0;\gamma) + \mathcal{N}(z; 0,\sigma \rightarrow 0)p(\pi=1;\gamma)\therefore\\
      &(1-\gamma)\mathcal{N}(z; 0, 1) + \gamma \mathcal{N}(z; 0,\sigma \rightarrow 0) \therefore
\end{aligned}$%
}
\end{equation*}
Therefore,
\begin{equation*}
\resizebox{.9\linewidth}{!}{$p(z;\gamma) = (1-\gamma)\mathcal{N}(z; 0, 1) + \gamma \mathcal{N}(z; 0,\sigma \rightarrow 0)$}.
\end{equation*}

\section{End-to-end Differentiable}
Sampling a value from the Spike-and-Slab posterior distribution $q(z|x, \gamma)$ is a two step process. First  a spike or slab component is sampled which is a binary decision,  we use Binary Concrete distribution \cite{DBLP:journals/corr/MaddisonMT16} to make this sampling step end-to-end differentiable. Then the value is sampled from the corresponding component, for this we employ the reparameterization trick \citep{DBLP:journals/corr/KingmaW13}. Also, samples from the Beta distribution are pathwise differentiable \citep{10.5555/3326943.3326984}.

\section{Hoyer}
\label{appendix:hoyer}
This section reports Average Hoyer, for the two corpora Yelp and Yahoo, both on the mean and samples from the posterior distributions of the HSVAE and MAT-VAE models.
\subsection{MAT-VAE}
\begin{figure}[H]%
\centering
\includegraphics[scale=0.205,trim={0cm 0.5cm 0cm 0}, clip]{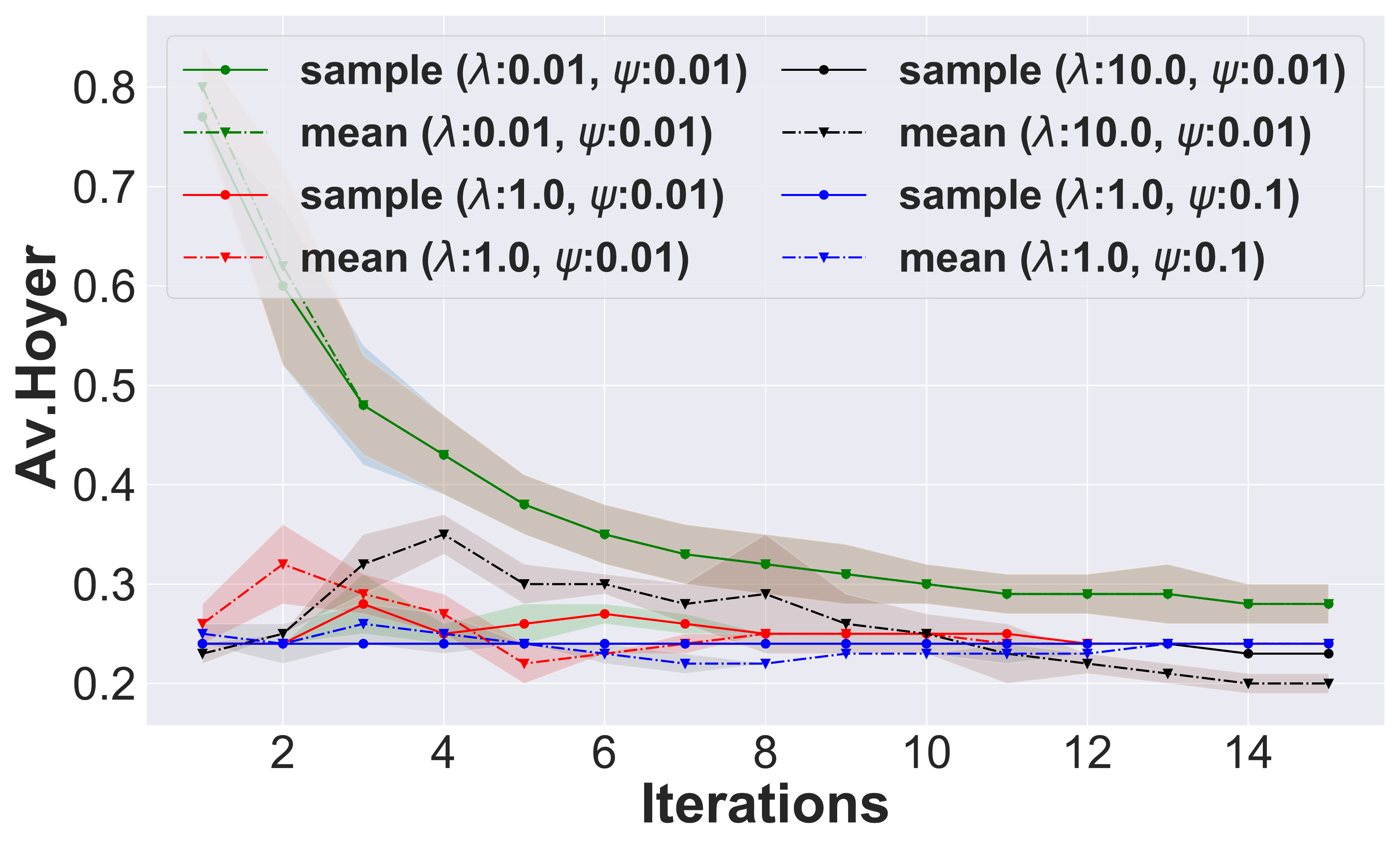}
\caption{Average Hoyer (Av.Hoyer) on Yelp corpus dev set  for MAT-VAE. Lines are an average over the 3 runs of the models, the shaded area is the standard deviation. The dimensionality of the latent variable of the models is 32D.}
\label{figure:hoyer}
\vspace{-2mm}
\end{figure}

\begin{figure}[H]%

\centering
\includegraphics[scale=0.205,trim={0cm 0.5cm 0cm 0}, clip]{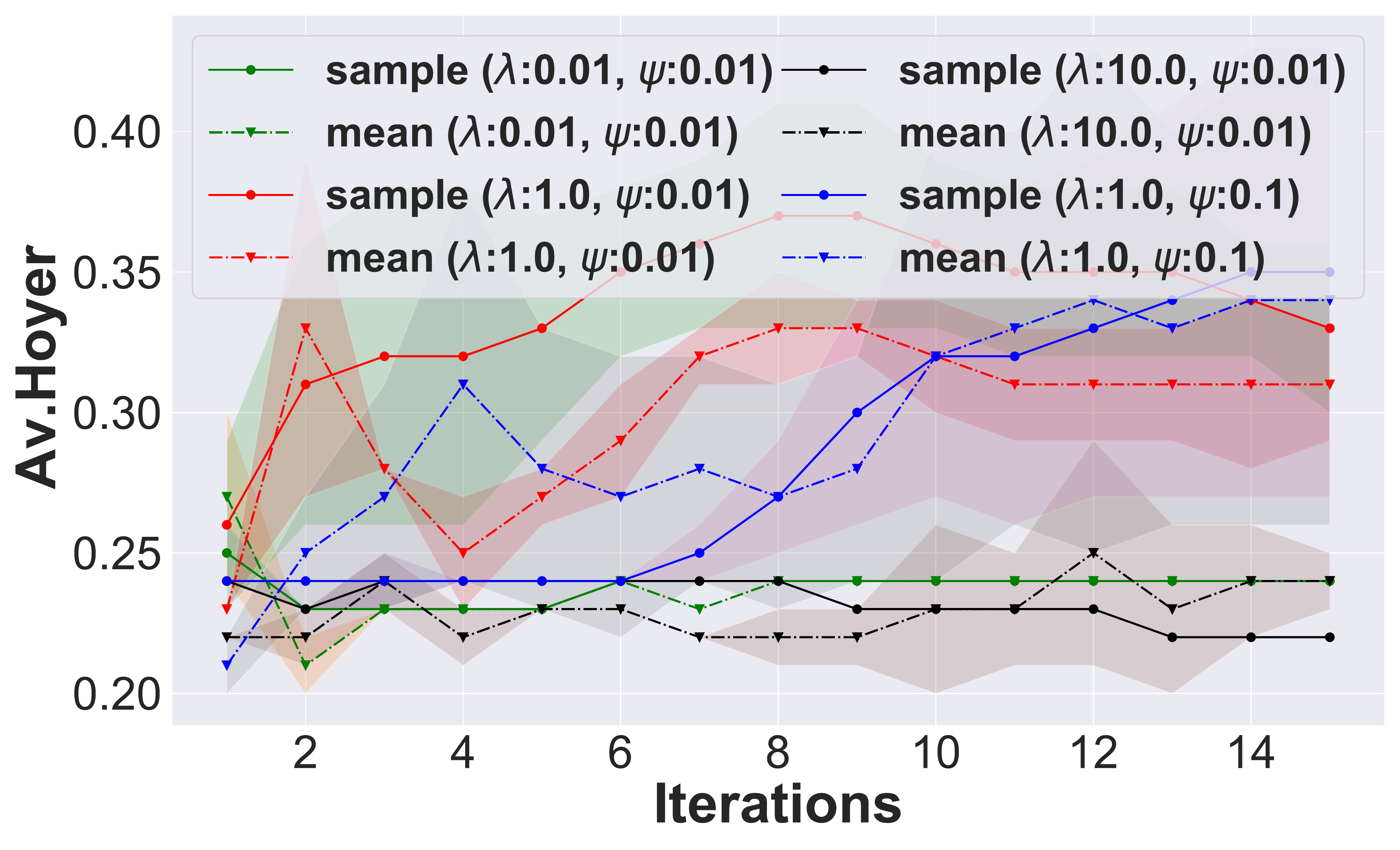}

\caption{Average Hoyer (Av.Hoyer) on Yahoo corpus dev set  for MAT-VAE.  Lines are an average over the 3 runs of the models, the shaded area is the standard deviation.}

\label{figure:hoyer}
\vspace{-2mm}
\end{figure}
\subsection{HSVAE}

\begin{figure}[H]%
\centering
\includegraphics[scale=0.205,trim={0cm 0.5cm 0cm 0}, clip]{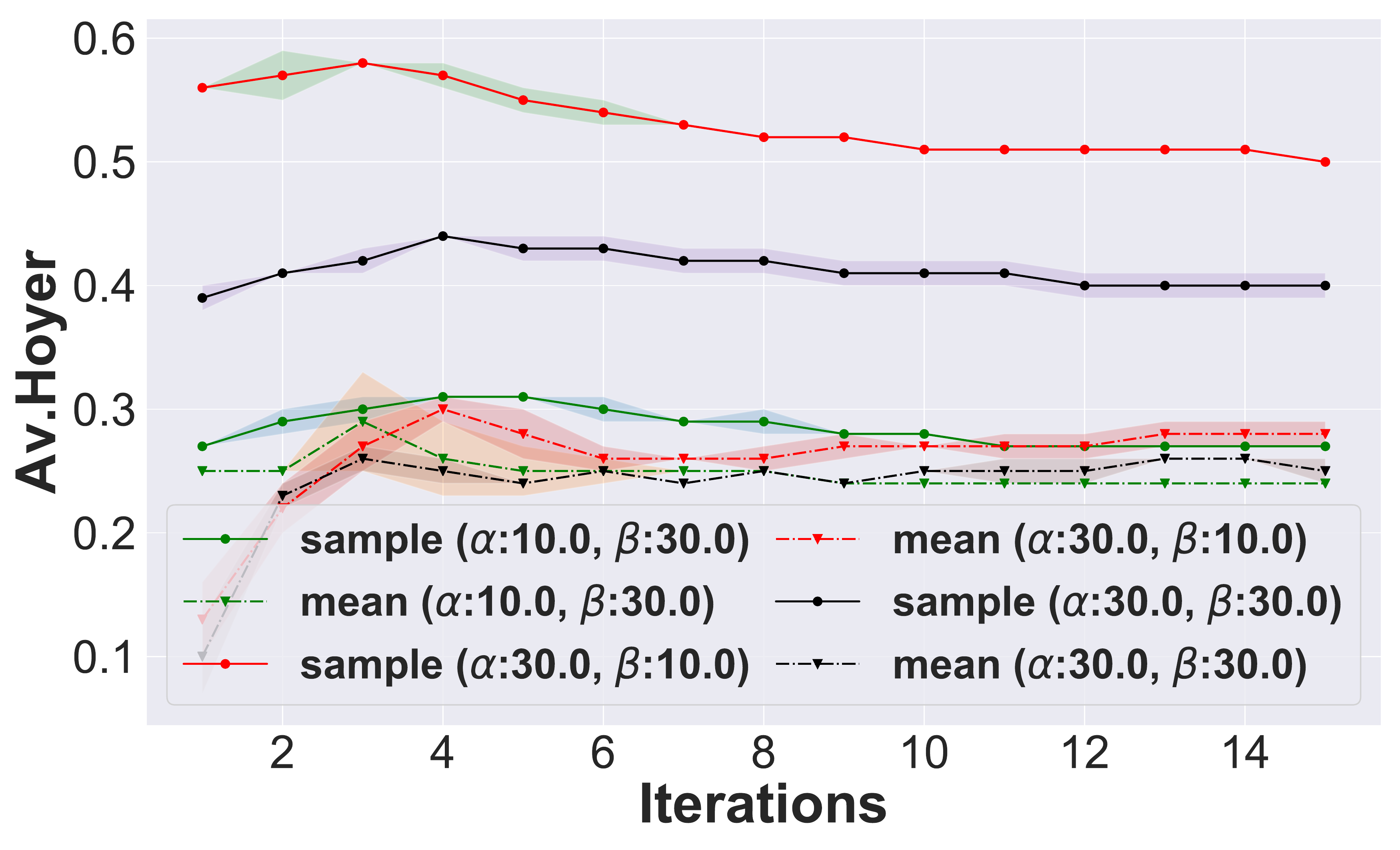}
\caption{Average Hoyer (Av.Hoyer) on Yelp corpus dev set  for HSVAE.  Lines are an average over the 3 runs of the models, the shaded area is the standard deviation. The dimensionality of the latent variable of the models is 32D.}
\label{figure:hoyer_3}
\vspace{-2mm}
\end{figure}

\begin{figure}[H]%

\centering
\includegraphics[scale=0.205,trim={0cm 0.5cm 0cm 0}, clip]{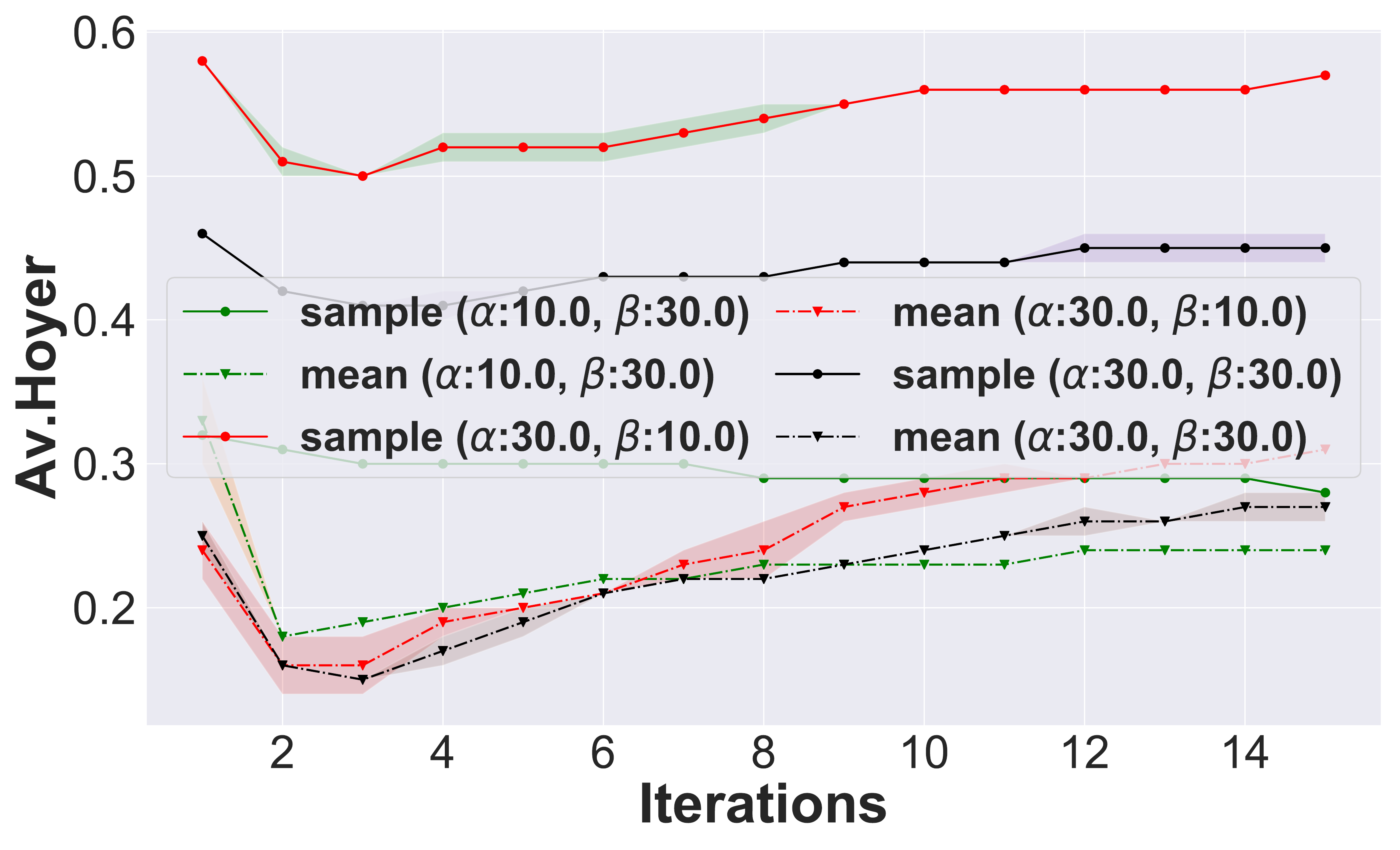}

\caption{Average Hoyer (Av.Hoyer) on Yahoo corpus dev set  for HSVAE.  Lines are an average over the 3 runs of the models, the shaded area is the standard deviation. The dimensionality of the latent variable of the models is 32D.}

\label{figure:hoyer_4}
\vspace{-2mm}
\end{figure}

\section{Hardware}
Please refer to  Table \ref{table:hardware} for the hardware that we use.
\begin{table}[H]
\setlength{\tabcolsep}{2pt}
\centering
\scalebox{0.70}{
\begin{tabular}{l r  }
\toprule

  hardware & specification \\
 \midrule
    CPU  & Intel\textsuperscript{\textregistered} Xeon E5-2670V3, 12-cores, 24-threads \\
  GPU  & NVIDIA\textsuperscript{\textregistered} TITAN RTX\textsuperscript{TM} (24 GB) x 1 \\
  RAM  & CORSAIR\textsuperscript{\textregistered}  Vengeance LPX  DDR4 2400 MHz (8 GB) x 4 \\

 \bottomrule
\end{tabular}
}
\caption{Computing infrastructure.}
\label{table:hardware}
\end{table}

\section{Datasets}

\begin{table}[H]
\setlength{\tabcolsep}{3pt}
\centering
\scalebox{0.70}{
\begin{tabular}{l c c c}
\toprule

 & \textbf{Yelp} & \textbf{DBpedia} & \textbf{Yahoo} \\
 \midrule

  \# sent. (train corpus)   & 100K & 140K & 100K \\
  \# sent. (valid corpus) & 10K & 14K& 10K \\
   \# sent. (test corpus)  & 10K & 14K& 10K \\
   vocabulary size  & 19,997 & 20K& 20K \\
   min sent. length.  & 20 & 1& 5 \\
   av. sent. length.  & 96 & 35& 12 \\
   max. sent. length. & 200 & 60& 30\\
   \# classes & 5 & 14& 10\\
   \# sent. in each class (train/test corpus)  & 20K/2K & 10K/1K& 10K/1K\\

 \bottomrule
\end{tabular}
}
\caption{Statistics of corpora. Vocabulary size excludes the \textlangle pad \textrangle and \textlangle EOS \textrangle symbols.}
\label{table:corpora_stat}
\end{table}

\end{document}